\def\year{2020}\relax
\documentclass[letterpaper]{article} 
\usepackage{aaai20}  
\usepackage{times}  
\usepackage{helvet} 
\usepackage{courier}  
\usepackage[hyphens]{url}  
\usepackage{graphicx} 
\usepackage{color}
\usepackage{bm}
\usepackage{url}
\usepackage{multirow}
\usepackage{booktabs}
\usepackage{amsmath, amssymb}

\newcommand{\newcite}[1]{\citeauthor{#1}~(\citeyear{#1})}
\def\argmax{\mathop{\rm argmax}}

\usepackage{graphicx}  
\title{Visual Agreement Regularized Training for Multi-Modal Machine Translation}
\author{Pengcheng Yang,\textsuperscript{\rm 1,\rm 2}
Boxing Chen,\textsuperscript{\rm 3}
Pei Zhang,\textsuperscript{\rm 3}
Xu Sun,\textsuperscript{\rm 1,\rm 2}
\\ 
\textsuperscript{\rm 1}{Center for Data Science , Peking University}\\
\textsuperscript{\rm 2}{MOE Key Lab of Computational Linguistics, School of EECS, Peking University}\\
\textsuperscript{\rm 3}{Alibaba DAMO Academy, Hangzhou, China}\\
\{yang\_pc, xusun\}@pku.edu.cn,
\{boxing.cbx, xiaoyi.zp\}@alibaba-inc.com
}

\begin{document}

\maketitle

\begin{abstract}
Multi-modal machine translation aims at translating the source sentence into a different language in the presence of the paired image. Previous work suggests that additional visual information only provides dispensable help to translation, which is needed in several very special cases such as translating ambiguous words. To make better use of visual information, this work presents visual agreement regularized training. The proposed approach jointly trains the source-to-target and target-to-source translation models and encourages them to share the same focus on the visual information when generating semantically equivalent visual words (e.g. \emph{``ball''} in English and \emph{``ballon''} in French). Besides, a simple yet effective multi-head co-attention model is also introduced to capture interactions between visual and textual features. The results show that our approaches can outperform competitive baselines by a large margin on the Multi30k dataset. Further analysis demonstrates that the proposed regularized training can effectively improve the agreement of attention on the image, leading to better use of visual information.

\end{abstract}

\section{Introduction}
As real-scenarios integrating multiple modal information have become commonplace, an increasing of endeavors have been paid to multi-modal machine translation (MMT). Different from tradition machine translation based solely on textual information, the MMT task aims at translating sentences paired with images into a different target language~\cite{elliott2016multi30k}. It not only has plenty of practical applications, e.g., helping improve the translation of ambiguous multi-sense words, but also serves as an ideal testbed for cross-modal text generation~\cite{calixto2017doubly}.

\begin{figure}
    \centering
    \includegraphics[width=1.0\linewidth]{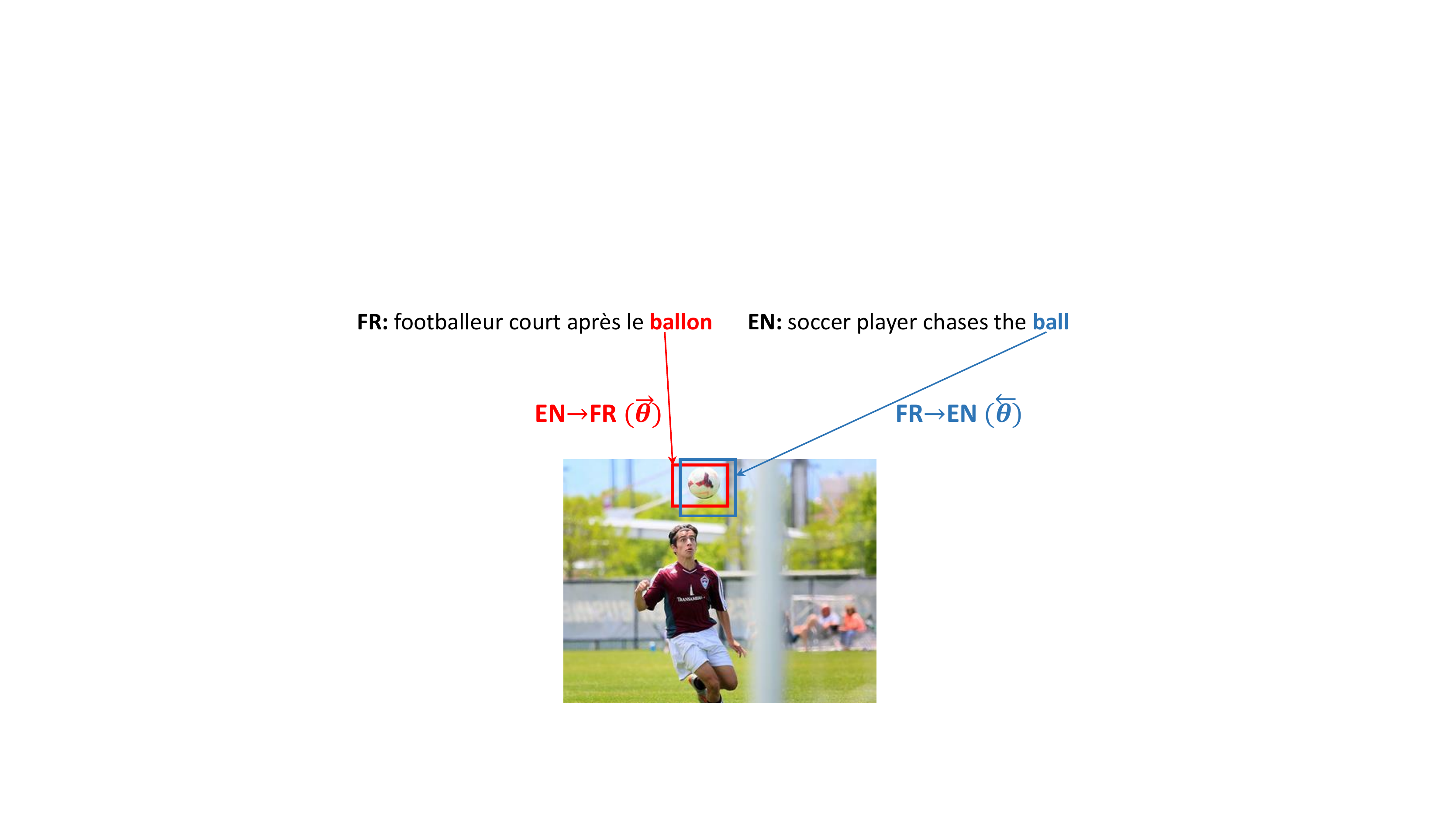}
    \caption{A toy example of visual agreement regularized training. $\overrightarrow{\bm{\theta}}$ denotes forward model translating English (EN) into French (FR) and $\overleftarrow{\bm{\theta}}$ is the opposite. When $\overrightarrow{\bm{\theta}}$ generates ``\emph{ballon}'' and $\overleftarrow{\bm{\theta}}$ generates ``\emph{ball}'', they should all focus on the region near the football in the image.}
    \label{fig:case}
\end{figure}

Despite its importance described above, previous work on MMT typically suffers from two drawbacks. First of all, how to effectively integrate visual information still remains an intractable challenge. Previous work has shown that additional visual information only provides dispensable help, which may be needed in the presence of some special cases (e.g. translating incorrect or ambiguous source words or generating gender-neutral target words). As a result, most existing approaches tend to ignore such visual information. 
To remedy this, we present visual agreement regularized training for the MMT task in this work. The proposed training schema jointly trains both source-to-target (forward) and target-to-source (backward) translation models, and encourages them to share the same focus on the image when generating semantic equivalent visual words (e.g. \emph{``ball''} in English and \emph{``ballon''} in French). Take Figure~\ref{fig:case} as an example, whose translation direction is EN$\to$FR. 
When generating the target word \emph{``ballon''}, the forward model is likely to focus on the red box region in the image, on which is also expected to be focused by the backward model when it generates the source word \emph{``ball''}.
By encouraging the agreement of bidirectional (forward and backward) models' attention (red and blue boxes in Figure~\ref{fig:case}) to image, the visual information can be more fully utilized.

However, a tricky problem is that semantically equivalent word pairs tend to be unavailable in practice. To address this issue, we present two effective regularization approaches: hard regularization constructing pseudo-aligned word pairs via external word aligner~\cite{fast_align}, and soft regularization employing posterior estimation to provide an approximation.
Considering that the generation of non-visual words (e.g., \emph{``the''} and \emph{``le''} in Figure~\ref{fig:case}) requires little or no visual information from the image, we further propose an adaptive weighting strategy to adaptively adjust the weight of regularization item based on the dependence of the word to be generated on the image. 

The other challenge is how to capture interactions between visual and textual features. Most previous work independently obtains representations of the input image and source sentence, which ignores the complex dependency between two information sources. To tackle this challenge, we introduce a simple yet effective multi-head co-attention, which can build bidirectional interactions between visual and textual features in multiple subspaces so that two information sources can mutually boost for better representations. 

The main contributions of this work are listed as follows:

\begin{itemize}
    \item In terms of model training, we propose two visual agreement regularized training schemas as well as an adaptive weighting strategy, which encourage bidirectional models to make better use of visual information.
    \item In terms of model architecture, we introduce a multi-head co-attention model. It aims to capture the interaction between visual and textual features in multiple subspaces so that two information sources can mutually boost for better representations.
    \item Experimental results show that our approach can outperform competitive baselines by a large margin. Further analysis demonstrates that bidirectional models can more fully utilize visual information by improving the agreement of visual attention.
\end{itemize}

\section{Proposed Model}
\label{sec:dca}

\begin{figure}
    \centering
    \includegraphics[width=1\linewidth]{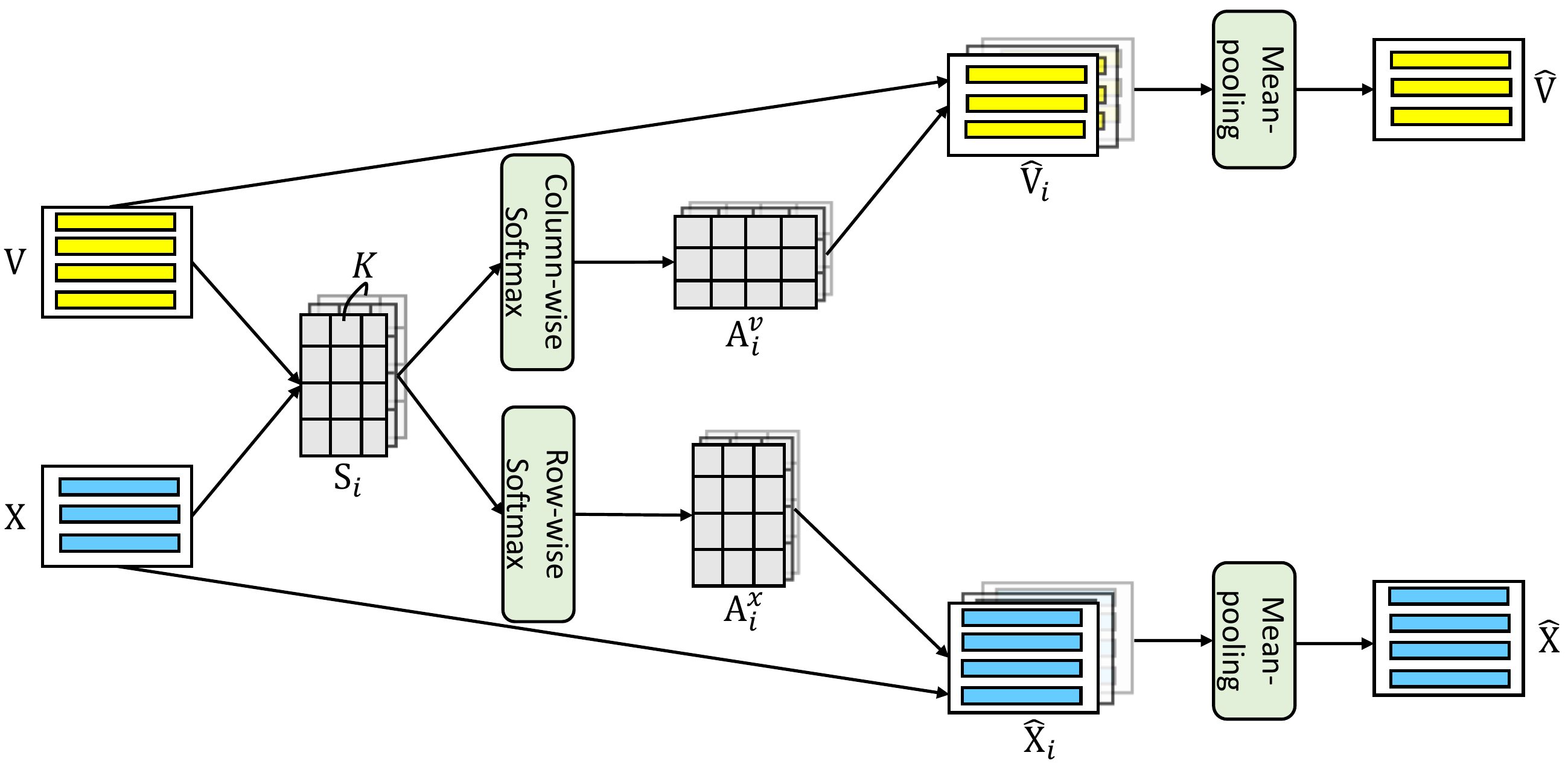}
    \caption{The illustration of multi-head co-attention model.}
    \label{fig:model}
\end{figure}

Given the visual features $\mathbf{v}=(v_1,\cdots,v_m)$ and source sentence $\mathbf{x}=(x_1,\cdots,x_n)$, the MMT task aims to generate the target sentence $\mathbf{y}=(y_1,\cdots,y_l)$. We first illustrate our approach with Seq2Seq model as base architecture and further extend it to Transformer. Figure~\ref{fig:model} presents the sketch of our proposed model, which is elaborated on as follows.

\subsection{Visual Encoder and Textual Encoder}
\label{sec:encoders}
The textual encoder is implemented as a recurrent neural network (RNN), which encodes the source sentence $\mathbf{x}$ into a sequence of hidden states. Formally, the hidden representation of each word $x_i$ is computed as $h_i^x = {\rm RNN}\big(h_{i-1}^x, e(x_i)\big)$, where $e(x_i)$ denotes the embedding of $x_i$. The final textual representation matrix is denoted as $\mathbf{X}=\{h_1^x,\cdots,h_n^x\}\in\mathbb{R}^{n \times d}$, where $n$ is the total number of textual representations and $d$ is the dimension of $h_i^x$.

The visual features $\mathbf{v}$ is pooled ROI features extracted by Faster R-CNN~\cite{ren2015faster}. 
The final visual representation matrix is denoted as $\mathbf{V}=\{h_1^v,\cdots,h_m^v\}\in\mathbb{R}^{m \times d}$, where $m$ is the total number of visual features.\footnote{Here we assume that $h_i^v$ and $h_i^x$ share the same dimension. Otherwise, a linear transformation can be introduced to ensure that the dimensions of both are the same.}

\subsection{Multi-Head Co-Attention}
\label{sec:dca_modules}

To effectively capture the complex dependency between visual and textual features, here we introduce a simple yet effective multi-head co-attention model, which aims to build bidirectional interactions between two information sources in multiple subspaces.
In the $k$-th subspace, following~\newcite{lu2016coatten}, we first connect visual representations $\mathbf{V}$ and textual representations $\mathbf{X}$ by computing the similarity matrix $\mathbf{S}_k \in \mathbb{R}^{m \times n}$ between them as follows:
\begin{equation}
\label{eq:similarity}
\mathbf{S}_k = \mathbf{V}\mathbf{M}_k\mathbf{X}^{\rm T}
\end{equation}
where $\mathbf{M}_k\in\mathbb{R}^{d\times d}$ is a trainable parameter matrix. The $(i,j)$ element of $\mathbf{S}_k$ in Eq.~(\ref{eq:similarity}) denotes the similarity between the $i$-th visual feature and the $j$-th textual feature. It is normalized row-wise to produce the image-to-text attention weights $\mathbf{A}_k^x$, and column-wise to produce the text-to-image attention weights $\mathbf{A}_k^v$:
\begin{equation}
\label{eq:attention}
     \mathbf{A}_k^x = {\rm softmax}(\mathbf{S}_k),\quad
     \mathbf{A}_k^v = {\rm softmax}(\mathbf{S}_k^{\rm T}) 
\end{equation}
where ${\rm softmax}(\cdot)$ means row-wise normalization. Further, we can obtain the image-aware textual representations $\widehat{\mathbf{X}}_k\in\mathbb{R}^{m\times d}$ and text-aware visual representations $\widehat{\mathbf{V}}_k\in\mathbb{R}^{n\times d}$ by multiplying attention weights with corresponding representations:
\begin{equation}
\label{eq:xhat}
     \widehat{\mathbf{X}}_k = \mathbf{A}_k^x\mathbf{X},\quad
     \widehat{\mathbf{V}}_k = \mathbf{A}_k^v\mathbf{V}
\end{equation}
In the calculation of $\widehat{\mathbf{X}}_k$ and $\widehat{\mathbf{V}}_k$, $\mathbf{X}$ and $\mathbf{V}$ mutually guide each other's attention. Therefore, these two sources of information can mutually boost for better representations.

To capture the interaction between two information sources in multiple subspaces, we set $K$ different parameter matrices $\{\mathbf{M}_1,\cdots,\mathbf{M}_K\}$, where $K$ is a hyper-parameter denoting the number of subspaces. 
Each $\mathbf{M}_k$ can yield a set of co-dependent representations $\widehat{\mathbf{X}}_k$ and $\widehat{\mathbf{V}}_k$ in this subspace according to Eq.~(\ref{eq:similarity}) - Eq.~(\ref{eq:xhat}). Finally, a mean-pooling layer is used to integrate all co-dependent representations from different subspaces:
\begin{equation}
\label{eq:mean_pooling}
     \widehat{\mathbf{X}} = \frac{1}{K}\sum_{k=1}^K\widehat{\mathbf{X}}_k,\quad
     \widehat{\mathbf{V}} = \frac{1}{K}\sum_{k=1}^K\widehat{\mathbf{V}}_k
\end{equation}

\subsection{Decoder}

The decoder implemented as another RNN model is responsible for generating target sentence $\mathbf{y}$. The hidden state $s_{t}$ of the decoder at time-step $t$ is computed as:
\begin{equation}
\label{eq:decoder_new}
s_{t} = {\rm RNN}\big(s_{t-1},\left[e(y_{t-1});\bar{c}_t^v;\bar{c}_t^x\right]\big)
\end{equation}
where $y_{t-1}$ is the word generated at time-step $t-1$. $\bar{c}_t^v$ and $\bar{c}_t^x$ are time-dependent visual and textual context vectors, respectively. Both $\bar{c}_t^v$ and $\bar{c}_t^x$ are obtained via attention mechanism and adaptive fusion, elaborated on as follows.

First, we compute an alignment score ${\rm A}_{y_t \to v_i}$ between each visual representation $h_i^v$ and the target word $y_t$ to be generated at the current time-step $t$ as:
\begin{equation}
\label{eq:visual_attn} 
{\rm A}_{y_t \to v_i} = \frac{\exp\big(a(s_{t-1}, h_i^v)\big)}{\sum_{j=1}^{m}\exp\big(a(s_{t-1}, h_j^v)\big)} 
\end{equation}
where $a(s_{t-1}, h_i^v)$ is an attention model measuring the dependency between $s_{t-1}$ and $h_i^v$. Readers can refer to~\newcite{seq2seq} for the details. Then, the preliminary visual context vector is obtained as:

\begin{equation}
c_t^v=\sum_{i=1}^m{\rm A}_{y_t \to v_i}h_i^v
\end{equation}
The preliminary textual context vector can be computed in a similar way:
\begin{align}
{\rm A}_{y_t \to x_i} &= \frac{\exp\big(a(s_{t-1}, h_i^x)\big)}{\sum_{j=1}^{n}\exp\big(a(s_{t-1}, h_j^x)\big)} \\
c_t^x&=\sum_{i=1}^n{\rm A}_{y_t \to x_i}h_i^x \label{equ5}
\end{align}
$\hat{c}_t^v$ and $\hat{c}_t^x$ can also obtained by attending to $\widehat{\mathbf{V}}$ and $\widehat{\mathbf{X}}$ with $s_{t-1}$ as query, respectively. We compute the final visual vector $\bar{c}_t^v$ by adaptively fusing ${c}_t^v$ and $\hat{c}_t^v$ as follows:
\begin{align}
g_t^v &= {\rm sigmoid}(\mathbf{U}_1{c}_t^v + \mathbf{U}_2\hat{c}_t^v) \\
\bar{c}_t^v &= g_t^v\odot{c}_t^v + (1-g_t^v)\odot\hat{c}_t^v
\end{align}
where $\mathbf{U}_1$ and $\mathbf{U}_2$ are trainable parameters. $\odot$ denotes element multiplication. The final textual context vector $\bar{c}_t^x$ can be obtained by adaptively fusing ${c}_t^x$ and $\hat{c}_t^x$ in a similar way.

\section{Visual Agreement Regularized Training}
In this section, we introduce our proposed visual agreement regularized training, which jointly trains both forward $P(\mathbf{y}|\mathbf{v},\mathbf{x};\overrightarrow{\bm{\theta}})$ and backward $P(\mathbf{x}|\mathbf{v},\mathbf{y};\overleftarrow{\bm{\theta}})$ translation models on the training corpus $\mathcal{D}=\{(\mathbf{v},\mathbf{x},\mathbf{y})\}$. Here $\overrightarrow{\bm{\theta}}$ and $\overleftarrow{\bm{\theta}}$ denote corresponding model parameters.

\subsection{Overview}
Given the instance $(\mathbf{v},\mathbf{x},\mathbf{y})$, the core idea of visual agreement regularized training is to encourage both forward and backward models to share the same focus on the image when generating semantically equivalent visual words. 
We use ${\rm A}_{y\to\mathbf{v}}(\overrightarrow{\bm{\theta}})=({\rm A}_{y\to v_1},\cdots,{\rm A}_{y\to v_m})$ to represent the attention vector of the forward model parameterized by $\overrightarrow{\bm{\theta}}$ on the image when it generates target word $y$. Similarly, ${\rm A}_{x^*\to\mathbf{v}}(\overleftarrow{\bm{\theta}})$ denotes the attention vector of the backward model parameterized by $\overleftarrow{\bm{\theta}}$ on the image when it generates source word $x^*$, where $x^* \in \mathbf{x}$ is the true aligned word of $y$. Then, the regularized training objective of forward model is defined as:
\begin{equation}
    \begin{split}
    L(\overrightarrow{\bm{\theta}}) =& {\rm log}P(\mathbf{y}|\mathbf{v},\mathbf{x};\overrightarrow{\bm{\theta}}) \\
    &-\lambda_1\sum_{y\in\mathbf{y}} \Delta\big({\rm A}_{y\to\mathbf{v}}(\overrightarrow{\bm{\theta}}),{\rm A}_{x^*\to\mathbf{v}}(\overleftarrow{\bm{\theta}})\big) \label{eq:constant}
    \end{split}
\end{equation}
where the regularization item $\Delta\big({\rm A}_{y\to\mathbf{v}}(\overrightarrow{\bm{\theta}}),{\rm A}_{x^*\to\mathbf{v}}(\overleftarrow{\bm{\theta}})\big)$ characterizes the difference between two attention vectors ${\rm A}_{y\to\mathbf{v}}(\overrightarrow{\bm{\theta}})$ and ${\rm A}_{x^*\to\mathbf{v}}(\overleftarrow{\bm{\theta}})$. We define $\Delta(\cdot,\cdot)$ as the MSE loss. Similarly, the regularized training objective of the backward model is defined as:
\begin{equation}
    \begin{split}
    \label{eq:obj_bw}
    L(\overleftarrow{\bm{\theta}}) =& {\rm log}P(\mathbf{x}|\mathbf{v},\mathbf{y};\overleftarrow{\bm{\theta}}) \\
    &-\lambda_2\sum_{x\in\mathbf{x}} \Delta\big({\rm A}_{x\to\mathbf{v}}(\overleftarrow{\bm{\theta}}),{\rm A}_{y^*\to\mathbf{v}}(\overrightarrow{\bm{\theta}})\big)
    \end{split}
\end{equation}

\subsection{Visual Agreement Regularization}
However, a tricky problem is that true aligned word pair $(y,x^*)$ or $(x,y^*)$ tends to be unavailable in practice. To address this issue, here we propose two solutions: hard and soft visual agreement regularization. For simplicity, we illustrate these two approaches based on the forward model. 

\paragraph{Hard visual agreement regularization.}
For each training instance $(\mathbf{v},\mathbf{x},\mathbf{y})$, the hard visual agreement regularization aims to align each target word $y \in \mathbf{y}$ with a source word $\hat{x} \in \mathbf{x}$ that has the highest alignment probability. Formally, 
\begin{equation}
x^*\approx\hat{x} = \argmax_{x\in\mathbf{x}}{\rm aligner}(x|y)
\end{equation}
where ${\rm aligner}(\cdot|\cdot)$ denotes the alignment probability that can be obtained by unsupervised word aligner.

\paragraph{Soft visual agreement regularization.}
For some target word $y$, there may be multiple source words to be aligned with it. For instance, in Figure~\ref{fig:soft_var}, the true translation of target word ``\emph{footballeur}'' is ``\emph{soccer player}'' consisting of two words. However, the hard regularization can only align one source word with $y$. To tackle this problem, here we propose soft visual agreement regularization, which aims at applying posterior estimation to provide an approximation of ${\rm A}_{x^*\to\mathbf{v}}(\overleftarrow{\bm{\theta}})$. Formally, we rewrite ${\rm A}_{x^*\to\mathbf{v}}(\overleftarrow{\bm{\theta}})$ as:
\begin{equation}
    \begin{split}
    \label{eq:soft}
    {\rm A}_{x^*\to \mathbf{v}}(\overleftarrow{\bm{\theta}}) &= \sum\limits_{x\in \mathbf{x}}\mathbb{I}(x=x^*){\rm A}_{x\to \mathbf{v}}(\overleftarrow{\bm{\theta}}) \\
    &= \mathbb{E}_{x\sim p^*(x|y)}{\rm A}_{x\to \mathbf{v}}(\overleftarrow{\bm{\theta}}) 
    \end{split}
\end{equation}
where $p^*(x|y)$ denotes the true aligned distribution concentrated\footnote{It means that $p^*(x|y)=1$ holds only when $x$ is $x^*$. } to the point $x^*$. 

For each target word $y\in\mathbf{y}$, its probability of being aligned with source word $x\in\mathbf{x}$ can be characterized by the attention weight of forward model to source word $x$ when generating $y$. Therefore, we can use the attention distribution ${\rm A}_{y\to\mathbf{x}}$ of target word $y$ to source sentence $\mathbf{x}$ to approximate $p^*(x|y)$ in Eq.~(\ref{eq:soft}), yielding the following estimation:
\begin{equation}
    \begin{split}
    {\rm A}_{x^*\to \mathbf{v}}(\overleftarrow{\bm{\theta}}) & \approx \mathbb{E}_{x\sim p(x|y,\overrightarrow{\bm{\theta}})}{\rm A}_{x\to \mathbf{v}}(\overleftarrow{\bm{\theta}})   \\
    &= \sum\limits_{x\in \mathbf{x}}p(x|y,\overrightarrow{\bm{\theta}}){\rm A}_{x\to \mathbf{v}}(\overleftarrow{\bm{\theta}}) \\
    &= \sum\limits_{x\in \mathbf{x}}{\rm A}_{y\to x}(\overrightarrow{\bm{\theta}}){\rm A}_{x\to \mathbf{v}}(\overleftarrow{\bm{\theta}})
    \end{split}
\end{equation}
where $p(x|y,\overrightarrow{\bm{\theta}})={\rm A}_{y\to x}(\overrightarrow{\bm{\theta}})$ is a posteriori estimation of the true alignment distribution $p^*(x|y)$.

\subsection{Adaptive Weighting}
\begin{figure}
    \centering
    \includegraphics[width=1.0\linewidth]{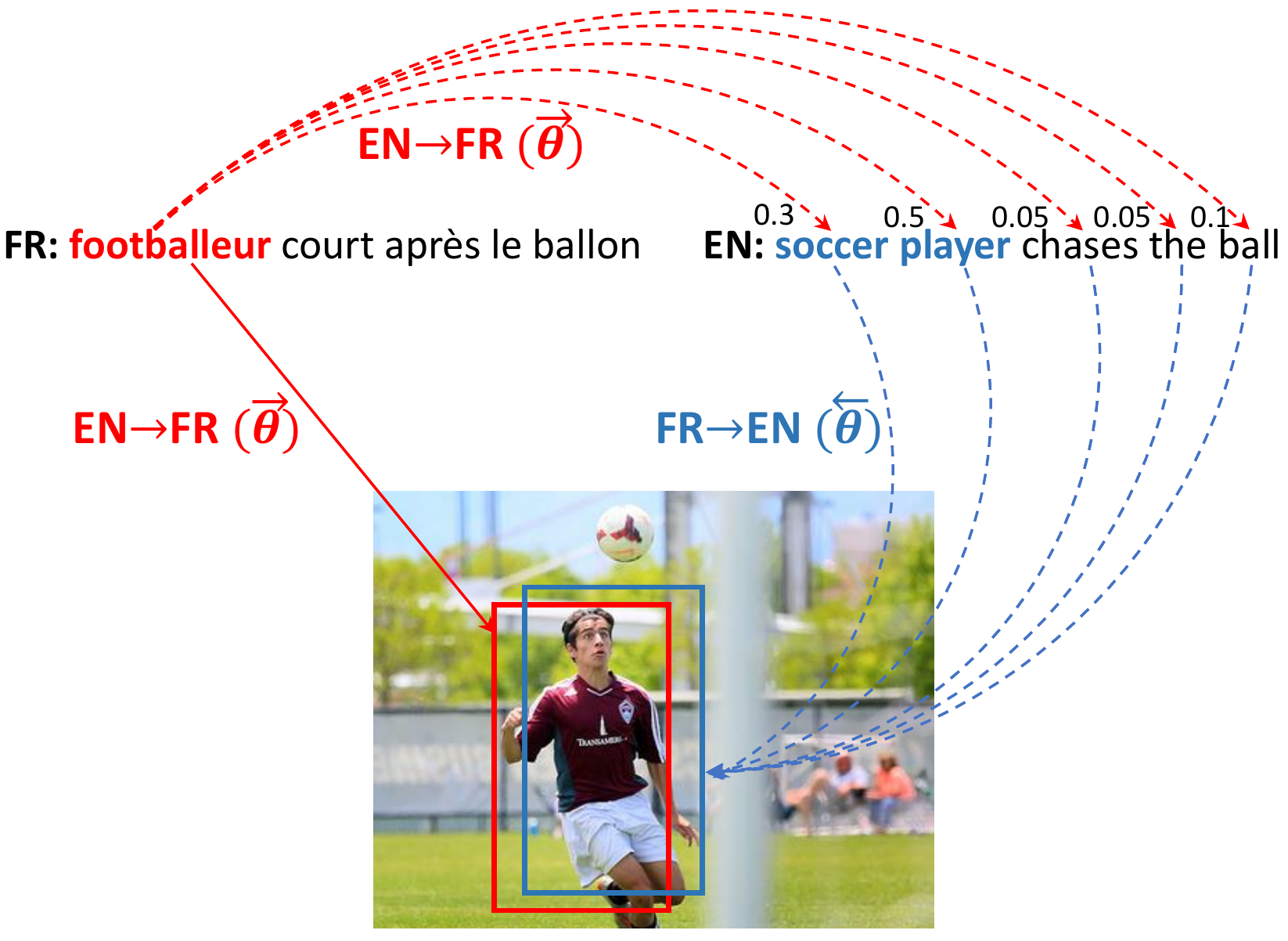}
    \caption{A toy example of soft regularization. The probability value 0.3 that ``\emph{footballeur}'' is aligned with ``\emph{soccer}'' can be defined as the attention weight of ``\emph{soccer}'' when the forward model parameterized by $\overrightarrow{\bm{\theta}}$ generates ``\emph{footballeur}''.}
    \label{fig:soft_var}
\end{figure}

During the translation process, there usually exist some non-visual words (e.g. ``\emph{le}'' and ``\emph{the}'' in Figure~\ref{fig:soft_var}) whose generation requires little or no visual information from the image, leading to no attention on the input image. Therefore, it is unreasonable to forcibly impose visual agreement regularization on such non-visual words, otherwise it will yield misleading gradient signal. To achieve this, we propose an effective adaptive weighting strategy, which equips the model with the capability of automatically determining whether to introduce visual agreement regularization based on the dependence of the word to be generated on the image.

In more detail, we employ gating mechanism to measure the expected importance of visual context vector $\bar{c}_t^v$ in relation to the next target word $y_t$ at time-step $t$ as follows:
\begin{align}
\beta_{y_t} & = {\rm sigmoid}\Big(w^{\rm T}{\rm tanh}(\mathbf{W}_1s_{t-1} + \mathbf{W}_2\bar{c}_t^v)\Big)
\end{align}
where $w, \mathbf{W}_1, \mathbf{W}_2$ are trainable parameters. We replace original visual vector $\bar{c}_t^v$ in Eq.~(\ref{eq:decoder_new}) with the rescaled visual vector $\beta_{y_t}\bar{c}_t^v$ to update the hidden state of decoder:
\begin{equation}
\label{eq:decoder2}
s_{t} = {\rm RNN}\big(s_{t-1},\left[e(y_{t-1});\beta_{y_t}\bar{c}_t^v;\bar{c}_t^x\right]\big)
\end{equation}

The gating scalar $\beta_{y_t}$ characterizes the dependence of the generation of $y_t$ on visual information. For visual words, the corresponding $\beta_{y_t}$ presents a larger value. Therefore, the new training objective can be modified to:
\begin{equation}
    \begin{split}
    L(\overrightarrow{\bm{\theta}}) =& {\rm log}P(\mathbf{y}|\mathbf{v},\mathbf{x};\overrightarrow{\bm{\theta}}) \\
    &-\lambda_1\sum_{y\in\mathbf{y}}\beta_{y}\Delta({\rm A}_{y\to\mathbf{v}}(\overrightarrow{\bm{\theta}}),{\rm A}_{x^*\to\mathbf{v}}(\overleftarrow{\bm{\theta}})) \label{eq:aw}
    \end{split}
\end{equation}

Different from the constant coefficient of regularization term in Eq.~(\ref{eq:constant}), we use $\beta_{y}$ to adaptively adjust the weight of regularization in Eq.~(\ref{eq:aw}). For visual words, $\beta_{y}$ presents a large value so that visual agreement regularization is introduced to make better use of visual information. Conversely, for non-visual words, the value of $\beta_{y}$ is small, avoiding the presence of misleading gradient signals caused by unreasonable regularization. Similarly, the training objective of the backward model can be modified to:
\begin{equation}
    \begin{split}
    \label{eq:adp_obj_bw}
    L(\overleftarrow{\bm{\theta}}) =& {\rm log}P(\mathbf{x}|\mathbf{v},\mathbf{y};\overleftarrow{\bm{\theta}}) \\
    &-\lambda_2\sum_{x\in\mathbf{x}}\beta_{x}\Delta({\rm A}_{x\to\mathbf{v}}(\overleftarrow{\bm{\theta}}),{\rm A}_{y^*\to\mathbf{v}}(\overrightarrow{\bm{\theta}}))
    \end{split}
\end{equation}

We also implement our multi-head co-attention model and visual agreement regularized training based on the Transformer~\cite{transformer}. The visual attention vector is calculated as the averaged attention from different heads of the decoder block which performs the best in word alignment. Due to the page limitation and this extension is not the focus of this paper, we will not explain it in more detail. We strongly recommend readers to refer to~\newcite{transformer} for more details.

\section{Experiments}
In this section, we introduce the dataset, evaluation metrics, all compared baselines as well as the detailed experiment settings.

\begin{table*}[!ht]
\begin{center}
\footnotesize
\begin{tabular}{lcccccccc}
\toprule
\multirow{2}{*}{Methods} & \multicolumn{2}{c}{EN$\to$DE} & \multicolumn{2}{c}{DE$\to$EN} & \multicolumn{2}{c}{EN$\to$FR} & \multicolumn{2}{c}{FR$\to$EN}\\ 
\cmidrule(r){2-3} \cmidrule(r){4-5} \cmidrule(r){6-7} \cmidrule(r){8-9}
&  BLEU  &  METEOR
&  BLEU  &  METEOR
&  BLEU  &  METEOR
&  BLEU  &  METEOR \\
\midrule
OnlyText & 27.1 & 49.3 & 34.2 & 33.6 & 51.3 & 69.1 & 46.8 & 40.7 \\
\midrule
DAMT & 27.5 & 49.4 & 34.8 & 34.1 & 51.2 & 68.9 & 47.1 & 41.1  \\
Imagination & 28.2 & 49.8 & 35.1 & 34.3 & 51.9 & 69.7 & 47.4 & 41.2 \\
\midrule
Proposal (Hard) & \textbf{29.3} & \textbf{51.2} & 35.5 & 35.1 & \textbf{52.6} & \textbf{69.9} & \textbf{48.9} & \textbf{41.6} \\
Proposal (Soft) & 29.2 & 50.9 & \textbf{35.7} & \textbf{35.2} & 52.4 & 69.5 & 48.7 & 41.5 \\
\bottomrule
\end{tabular}
\end{center}
\caption{The experimental results of Seq2Seq-based systems. The best performance is highlighted in bold.}
\label{tab:res_seq2seq}
\end{table*}

\begin{table*}[!ht]
\begin{center}
\footnotesize
\begin{tabular}{lcccccccc}
\toprule
\multirow{2}{*}{Methods} & \multicolumn{2}{c}{EN$\to$ DE} & \multicolumn{2}{c}{DE$\to$EN} & \multicolumn{2}{c}{EN$\to$FR} & \multicolumn{2}{c}{FR$\to$EN}\\ 
\cmidrule(r){2-3} \cmidrule(r){4-5} \cmidrule(r){6-7} \cmidrule(r){8-9}
&  BLEU  &  METEOR
&  BLEU  &  METEOR
&  BLEU  &  METEOR
&  BLEU  &  METEOR \\
\midrule
OnlyText & 27.9 & 49.1 & 32.2 & 32.4 & 52.5 & 69.9 & 45.9 & 40.2 \\
\midrule
DAMT & 28.3 & 49.7 & 32.1 & 32.2 & 52.2 & 70.0 & 46.1 & 40.1 \\
Imagination & 28.4 & 49.3 & 32.7 & 32.9 & 53.1 & 70.2 & 46.8 & 40.5 \\
\midrule
Proposal (Hard) & 29.3 & 50.2 & \textbf{33.4} & \textbf{33.5} & \textbf{53.3} & \textbf{70.4} & \textbf{47.7} & \textbf{40.8} \\
Proposal (Soft) & \textbf{29.5} & \textbf{50.3} & 32.9 & 33.1 & 52.8 & 70.1 & 47.3 & 40.6 \\
\bottomrule
\end{tabular}
\end{center}
\caption{The experimental results of Transformer-based systems. The best performance is highlighted in bold.}
\label{table:res_transformer}
\end{table*}

\subsection{Dataset}
Following previous work~\cite{calixto2017doubly}, we evaluate both our approach and all baselines on the Multi30K dataset~\cite{elliott2016multi30k}, which contains 29,000 instances for training and 1,014 for development. We use test-2017 for evaluation, which consists of 1,000 testing instances. Each image is paired with its English (EN) descriptions as well as human translations of German (DE) and French (FR). We use Moses SMT Toolkit to normalize and tokenize all sentences.

\subsection{Evaluation Metrics}
Following previous work~\cite{elliott2017imagination}, we adopt the following automatic evaluation metrics:

\begin{itemize}
    \item \textbf{BLEU}\footnote{\url{https://github.com/moses-smt/mosesdecoder/blob/master/scripts/generic/multi-bleu.perl}}~\cite{Bleu} compares the generated text with reference translations by computing overlapping lexical $n$-gram units.
    \item \textbf{METEOR}\footnote{\url{http://www.cs.cmu.edu/~alavie/METEOR/}}~\cite{Meteor} scores the generated text by aligning them to reference translations based on exact, stem, synonym, and paraphrase matches between words and phrases.
\end{itemize}

\subsection{Settings}
We utilize \emph{fast align}\footnote{\url{https://github.com/clab/fast_align}}~\cite{fast_align} to perform word alignment in the hard visual agreement regularization. Faster R-CNN are used to extract pooled ROI features as visual input for each region. For each image, we consistently keep 36 highest probability objects. $\lambda_1$ in Eq.~(\ref{eq:aw}) is set to 0.2 and 0.5 for EN$\to$DE and EN$\to$FR translations,respectively. $\lambda_2$ in Eq.~(\ref{eq:adp_obj_bw}) is set to 0.2 and 0.1 for DE$\to$EN and FR$\to$EN translations,respectively.

For both source and target language, we limit the vocabulary size to 10,000. The size of word embedding is set to 512 and embeddings are learned from scratch. An extra linear layer are utilized to project all visual features into 512-dim. For the Seq2Seq version of our approach, the textual encoder and decoder are all a 2-layer LSTM with hidden size 512. We set the textual encoder to be bidirectional. For the transformer version of our approach, we set the hidden size of multi-head attention layer to 512 and the hidden size of the feed-forward layer to 2,048. The number of heads in multi-head attention is set to 8, while a transformer layer consists of 6 blocks. 

We adopt the Adam optimization method with the initial learning rate 0.0003 for training and the learning rate is halved after each epoch. We also make use of dropout to avoid over-fitting. 

\subsection{Baselines}
We compare our proposed approaches with the following representative and competitive systems:

\begin{itemize}
    \item \textbf{OnlyText}~\cite{seq2seq} is the traditional sequence-to-sequence framework with attention mechanism, which only encodes the source sentence as the input.
     \item \textbf{DAMT}~\cite{calixto2017doubly} employs two separate attention mechanisms to integrate visual and textual features, respectively. Besides, the gate mechanism is further used to rescale visual information.
    \item \textbf{Imagination}~\cite{elliott2017imagination} learns both machine translation and visually grounding task simultaneously so that the vision-language joint semantic embeddings can be constructed.
\end{itemize}

\section{Results and Analysis}
In this section, we report the experimental results. Besides, further analysis is also provided.

\begin{table}
\begin{center}
\footnotesize
\begin{tabular}{lcccc}
\toprule
\multirow{2}{*}{Methods} & \multicolumn{2}{c}{EN$\to$DE} & \multicolumn{2}{c}{EN$\to$FR} \\ 
\cmidrule(r){2-3} \cmidrule(r){4-5} 
&  BLEU  &  METEOR
&  BLEU  &  METEOR \\
\midrule
OnlyText & 27.1 & 49.3 & 51.3 & 69.1 \\
\midrule
+ Image & 27.5 & 49.4 & 51.2 & 68.9   \\
+ MHCA & 28.6 & 50.1 & 51.7 & 69.3   \\
\midrule
Proposal (Hard) & \textbf{29.3} & \textbf{51.2} & \textbf{52.6} & \textbf{69.9}  \\
Proposal (Soft) & 29.2 & 50.9 & 52.4 & 69.5 \\
\bottomrule
\end{tabular}
\end{center}
\caption{The results of incremental analysis. The best performance is highlighted in bold and ``MHCA'' means our proposed multi-head co-attention.}
\label{tab:abltion_study}
\end{table}

\subsection{Experimental Results}
\label{sec:results}

The evaluation results of different systems based on the Seq2Seq model are presented in Table~\ref{tab:res_seq2seq}, showing that both our hard and soft visual agreement regularization achieve better performance than competitive baselines in all translation directions.
For instance, our soft regularization model outperforms the text-only Seq2Seq model and the best-performing baseline by 2.2 and 1.1 BLEU score respectively on the EN$\to$DE language pair.
This illustrates that visual features are capable of promoting the performance of machine translation, and our approaches can make better use of visual information.

Table~\ref{table:res_transformer} presents the results of different systems based on Transformer, showing that both regularization approaches can substantially outperform the Transformer-based baselines. This demonstrates that our approaches are universal, which can bring about consistent improvements in performance on different base neural architectures. 
However, it is worth noting that although soft regularization can achieve better performance than all baselines in most translation directions, its performance is not comparable to hard regularization. 
The reason for this observation, as we suspect, might fall in the multi-head attention mechanism in Transformer, where word alignment is not well performed~\cite{li2019word}. This leads to larger error when leveraging attention distribution to approximate the true alignment distribution, and the overall performance would be correspondingly inferior.

\subsection{Incremental Analysis}

Table~\ref{tab:abltion_study} presents results of incremental analysis. We treat the text-only Seq2Seq model as the base model and cumulatively add each component until the full model is rebuilt. The results show that the multi-head co-attention model can significantly improve the model performance, illustrating that building bidirectional interactions between textual and visual features can produce better co-dependent representations. Besides, Table~\ref{tab:abltion_study} shows that either hard or soft regularization further contributes to generating high-quality translations. By constraining bidirectional models to focus on the same region of the image when generating semantically equivalent visual words, the model can make better use of visual information, resulting in more accurate machine translation hypotheses.

\subsection{Effectiveness of Adaptive Weighting}

Considering that the generation of non-visual words requires little visual information from the image, we propose adaptive weighting, which aims to assign reasonable weights to visual regularization term automatically. Table~\ref{tab:aw} shows the performance of soft-agreement regularization with two different weighting strategies. In Table~\ref{tab:aw}, ``Frozen-WT'' and ``Adaptive-WT'' denote frozen weighting\footnote{Frozen weighting means that different words are assigned the same constant weight, as in Eq.~(\ref{eq:constant}).} and adaptive weighting, respectively.

\begin{table}
\begin{center}
\footnotesize
\begin{tabular}{lcccc}
\toprule
\multirow{2}{*}{Methods} & \multicolumn{2}{c}{EN$\to$DE} & \multicolumn{2}{c}{EN$\to$FR} \\ 
\cmidrule(r){2-3} \cmidrule(r){4-5} 
&  BLEU  &  METEOR &  BLEU  &  METEOR \\
\midrule
OnlyText & 27.1 & 49.3 & 51.3 & 69.1 \\
\midrule
Frozen-WT  & 28.8 & 50.7 & 52.1 & 69.4  \\
Adaptive-WT & \textbf{29.2} & \textbf{50.9} & \textbf{52.4} & \textbf{69.5}  \\
\bottomrule
\end{tabular}
\end{center}
\caption{The results of two weighting (WT) strategies based on the soft regularization.}
\label{tab:aw}
\end{table}

As shown in Table~\ref{tab:aw}, adaptive weighting can achieve better performance, increasing BLEU score from 28.8 to 29.2 on the EN$\to$DE language pair. The reason is that based on the degree of dependence of the word to be generated on the image, the proposed adaptive weighting can adaptively adjust the weight of the regularization term. The weights corresponding to visual words are larger, which introduces visual agreement regularization to leverage visual information more effectively. For non-visual words, however, a smaller weight will be assigned to the regularization term in the training objective, which prevents the presence of misleading gradient signals.

\begin{figure}
    \centering
    \includegraphics[width=1\linewidth]{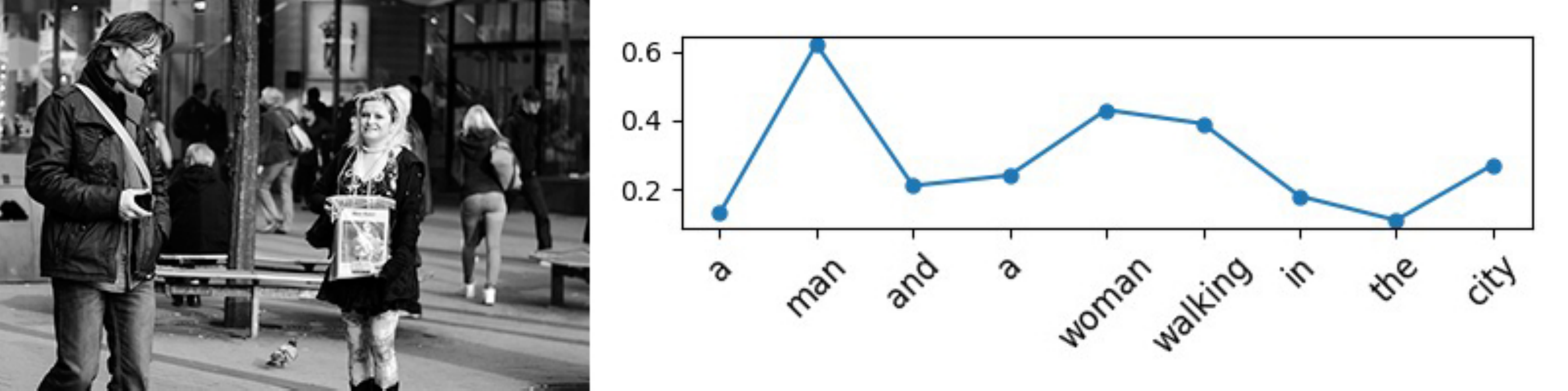}
    \caption{The visualization of the dependence of the right different words to be generated on the left image.}
    \label{fig:visual}
\end{figure}

Figure~\ref{fig:visual} visualizes the dependence (defined as $\beta_{y_{t}}$ in Eq.~(\ref{eq:decoder2})) of the different words to be generated on the image.
The results show that some visual words (such as \emph{``man''} and \emph{``woman''}) will be assigned greater weights, while non-visual words have a very small weight. It shows that the proposed adaptive weighting strategy can automatically adjust the corresponding weight according to the degree of dependence of the generated words on the image, resulting in better model performance.

\subsection{Effectiveness of Improving Visual Agreement}
\label{sec:vad}
In order to verify that our approach can effectively improve the agreement of visual attention of bidirectional models, here we introduce a new evaluation metric: visual agreement distance (\texttt{VAD}). The proposed \texttt{VAD} aims at characterizing the difference of visual attention when generating semantically equivalent words. Formally, for the given instance $(\mathbf{v},\mathbf{x},\mathbf{y})$, we define \texttt{VAD} as:
\begin{equation}
    \texttt{VAD} = \frac{1}{|\mathcal{N}(\mathbf{x},\mathbf{y})|} \sum\limits_{(x,y) \in \mathcal{N}(\mathbf{x},\mathbf{y})}\ell_1\Big({\rm A}_{x\to \mathbf{v}}, {\rm A}_{y\to \mathbf{v}}\Big) 
\end{equation}
where $\ell_1(\cdot,\cdot)$ denotes $\ell_1$ distance. $\mathcal{N}(\mathbf{x},\mathbf{y})$ denoted mutually aligned word pairs in $(\mathbf{x},\mathbf{y})$, which is defined as:
\begin{equation}
    \mathcal{N}(\mathbf{x},\mathbf{y}) = \Big\{(x,y) | x \in \mathbf{x}, y \in \mathbf{y}, x\leftrightarrow y\Big\} 
\end{equation}
where $x\leftrightarrow y$ represents that $x$ and $y$ are uniquely aligned with each other. The large \texttt{VAD} means that the model suffers from the poor visual agreement. The evaluation results of VAD for different systems on EN$\to$DE and EN$\to$FR translations are shown in Table~\ref{tab:vad}.

In Table~\ref{tab:vad}, \emph{non-visual} and \emph{visual} denote \texttt{VAD} calculated on non-visual word pairs and the remaining visual word pairs\footnote{We manually labeled such data for testing.}, respectively. 
The results show that our approach can effectively improve the agreement of visual attention to the image of bidirectional models in the generation of semantically equivalent word pairs. In addition, such improvement is more obvious for visual word pairs compared with non-visual word pairs (e.g. 0.44 decline for \emph{visual} v.s. 0.22 decline for \emph{non-visual} on EN$\to$DE). 
As analyzed in the previous section, the generation of visual words exhibits more dependence on the image. Therefore, the adaptive weighting will assign a greater weight to the corresponding regularization, leading to a more significant improvement.

\section{Related Work}

In summary, this work is mainly related to the following several research lines.

\paragraph{Multi-modal machine translation.} 
This task aims at translating the source sentence paired with additional modal (usually visual) information into another different language. 
Early work focuses on integrating additional visual features as part of inputs or the initialization of model modules. For instance, \newcite{huang2016attention} present parallel LSTM threads with multiple regional visual features and \newcite{calixto2017incorporating} propose to treat visual features as words in the source sentence or initialization of the encoder. Going a step further, several prior works employ separate attention for different modalities to integrate modal-dependent information. For example, \newcite{calixto2017doubly} present a doubly-attentive decoder integrating two separate attention over the source information and a more effective hierarchical attention was proposed by \newcite{DBLP:journals/corr/DelbrouckD17aa}. 
\newcite{ive2019distilling} propose an effective translation-and-refine framework, where visual features are only used by a second stage decoder. 
Inspired by multi-task learning, \newcite{elliott2017imagination} perform machine translation while constraining the averaged representations of the shared encoder to be the visual embedding of the paired image. \newcite{zhou2018vag} strives to construct a vision-language joint semantic embedding via a novel visual directional attention. 
There also exist some other endeavors on the MMT task. For example, \newcite{calixto2019latent} sets a latent variable as a stochastic embedding which is used in the target-language decoder and to predict visual features. \newcite{chen2019words} present a progressive learning approach for image pivoted zero-resource machine translation and \newcite{su2019unsupervised} investigate the possibility of utilizing images for disambiguation to improve the performance of unsupervised machine translation.

\begin{table}
\begin{center}
\footnotesize
\begin{tabular}{lcccc}
\toprule
\texttt{VAD} ($\downarrow$) & \multicolumn{2}{c}{EN$\to$DE} & \multicolumn{2}{c}{EN$\to$FR} \\ 
\cmidrule(r){1-1} \cmidrule(r){2-3} \cmidrule(r){4-5} 
Methods & \emph{visual} & \emph{non-visual} & \emph{visual} & \emph{non-visual} \\
\midrule
\emph{w/o agreement} & 1.18 & 1.71 & 1.13 & 1.52 \\
\emph{w/o agreement} & 0.74 & 1.49 & 0.71 & 1.45 \\
\bottomrule
\end{tabular}
\end{center}
\caption{The results of visual attention distance (\texttt{VAD}) on the EN$\to$DE and EN$\to$FR translations. The symbol ``$\downarrow$'' indicates that lower is better.}
\label{tab:vad}
\end{table}

\paragraph{Agreement-based machine translation.} 
Among this line of work, \newcite{cheng2016agreement} present agreement-based joint training for bidirectional neural machine translation systems and \newcite{liu2016agreement} propose to make use of approximate joint search for coupled translation models to produce more balanced outputs. Furthermore, \newcite{zhang2019regularize} present a model regularization approach by minimizing Kullback-Leibler divergence between the probability distributions defined by bidirectional models. Other representative translation tasks also include bilingual lexicon induction~\cite{artetxe2018robust}, unsupervised machine translation~\cite{lample2018phrase},  and so on.

\paragraph{Cross-modal generation.} 
This work can also be attributed to the category of cross-modal generation, which aims at generating the desired text in the presence of the source information from multiple modalities. Several typical tasks include image captioning~\cite{xu2015show,anderson2018bottom}, visual storytelling~\cite{wang2018no,yang2019visual},  cross-modal machine commenting~\cite{yang2019cross} and so on.

\section{Conclusion}
This work presents two novel visual agreement regularization approaches as well as an effective adaptive weighting strategy for multi-modal machine translation. By encouraging both forward and backward translation models to share the same focus on the image when generating semantically equivalent visual words, the proposed regularized training is capable of making better use of visual information. In addition, a simple yet effective multi-head co-attention model is also introduced to capture the interaction between visual and textual features. Extensive experimental results show that our approach can outperform competitive baselines by a large margin. Further analysis demonstrates that the proposed regularization approaches can effectively improve the agreement of attention on the image, leading to better use of visual information.

\section{Acknowledgement}
We thank all reviewers for their thoughtful and constructive suggestions. We also thank Kai Fan and Xuancheng Ren for their instructive suggestions and invaluable help. This work was supported in part by National Natural Science Foundation of China (No. 61673028). Xu Sun is the corresponding author of this paper.

\bibliography{aaai20}
\bibliographystyle{aaai}

\end{document}